\documentclass{article}

 \usepackage[preprint]{neurips_2025}

\usepackage[utf8]{inputenc} % allow utf-8 input
\usepackage[T1]{fontenc}    % use 8-bit T1 fonts
\usepackage{hyperref}       % hyperlinks
\usepackage{url}            % simple URL typesetting
\usepackage{booktabs}       % professional-quality tables
\usepackage{amsfonts}       % blackboard math symbols
\usepackage{nicefrac}       % compact symbols for 1/2, etc.
\usepackage{microtype}      % microtypography

\usepackage{color,colortbl}

\usepackage{multirow}
\usepackage{array}
\usepackage{graphicx}
\usepackage{amsmath}

\usepackage{caption}

\usepackage{multicol}

\definecolor{lightgray}{gray}{0.9}
\def\modelname{GeoVLA}
\def\ModuleActionExpert{3D-enhanced Action Expert}
\def\ModuleActionExpertAbbr{3DAE}
\def\ModulePointEncoder{Point Embedding Network}
\def\ModulePointEncoderAbbr{PEN}
\title{\modelname: Empowering 3D Representations in Vision-Language-Action Models}

\newcommand{\authorskip}{\hspace{4.8mm}}

\author{
Lin Sun$^{1}$\thanks{This work was done during the internship at Dexmal.}~~\thanks{Equal contribution.}
\quad Bin Xie$^{2}$\footnotemark[2]
\quad Yingfei Liu$^{2}$
\quad Hao Shi$^{3}$ 
\quad \textbf{Tiancai Wang}$^{2}$ 
\quad \textbf{Jiale Cao}$^{1}$\thanks{Corresponding author: Jiale Cao}\\[4pt]
$^1$Tianjin University \authorskip ~$^2$Dexmal \authorskip ~$^3$Tsinghua University\\[4pt]
\texttt{\{sun0806,connor\}@tju.edu.cn},\quad
\texttt{\{xiebin,lyf,wtc\}@dexmal.com}
}

\begin{document}

\maketitle

\maketitle

\begin{abstract}
  Vision-Language-Action (VLA) models have emerged as a promising approach for enabling robots to follow language instructions and predict corresponding actions. However, current VLA models mainly rely on 2D visual inputs, neglecting the rich geometric information in the 3D physical world, which limits their spatial awareness and adaptability. In this paper, we present \modelname, a novel VLA framework that effectively integrates 3D information to advance robotic manipulation. It uses a vision-language model (VLM) to process images and language instructions, extracting fused vision-language embeddings. In parallel, it converts depth maps into point clouds and employs a customized point encoder, called \ModulePointEncoder, to generate 3D geometric embeddings independently. These produced embeddings are then concatenated and processed by our proposed spatial-aware action expert, called \ModuleActionExpert, which combines information from different sensor modalities to produce precise action sequences. Through extensive experiments in both simulation and real-world environments, \modelname ~demonstrates superior performance and robustness. It achieves state-of-the-art results in the LIBERO and ManiSkill2 simulation benchmarks, and shows remarkable robustness in real-world tasks requiring height adaptability, scale awareness and viewpoint invariance. The project is available at \url{https://linsun449.github.io/GeoVLA}.
\end{abstract}    
\section{Introduction}
\label{sec:intro}

Advancing robot manipulation requires both intellectual interaction and precise physical motion control in real world.  Recently, vision-language-action (VLA) models, capable of instruction following and robotic action execution, have attracted significant attention. To leverage general knowledge, most VLA models are built upon vision-language models (VLMs)~\citep{liu2023visual, alayrac2022flamingo, chen2023pali, karamcheti2024prismatic,bai2023qwen} and develop through specialized designs for action generation. Early approaches such as RT-2~\citep{brohan2023rt} and OpenVLA~\citep{kim2024openvla} quantize the action space into discrete bins and adopt autoregressive token generation. While compatible with standard VLM architectures, this coarse representation struggles with complex, fine-grained manipulation tasks. To address this issue, some approaches~\citep{bjorck2025gr00t, black2024pi_0,li2024cogact, liu2025hybridvla} introduce specialized action experts that process features from VLMs and output action chunking \citep{zhao2023act} with diffusion processes~\citep{ho2020denoising} or flow matching~\citep{lipman2022flow} to directly parameterize continuous action spaces.

Despite these advancements, current VLA models predominantly rely on 2D visual inputs, overlooking the rich geometric priors inherent in 3D physical world. In contrast, 3D geometric information inherently provides accurate depth cues, enhanced spatial understanding, and robustness to viewpoint changes. Building on recent advances in 3D perception, emerging approaches such as LLaVA-3D~\citep{zhu2024llava} and SpatialVLA~\citep{qu2025spatialvla} integrate 3D positional encodings into VLMs to enable geometrically-aware representations. However, this integration disrupts the alignment between visual encoders and Large Language Models (LLMs). Consequently, bridging this misalignment often requires large-scale 3D embodied instruction-tuning datasets. As an alternative to such data-intensive approaches, another line of work focuses on injecting 3D information directly into action experts. For instance, 
PointVLA~\citep{li2025pointvla} adopts a two-stage training scheme: it first trains a 2D-VLA model, then freezes the action expert and injects point cloud features through a zero-initialized ControlNet-style module~\citep{zhang2023controlnet}. However, although freezing the action expert preserves its low-level capabilities, it hinders adaptation to the newly introduced point cloud modality. 
Therefore, integrating 3D information into VLA frameworks in an end-to-end manner remains a key challenge.

\begin{figure}[t]
    \centering
    \includegraphics[width=1.0\textwidth]{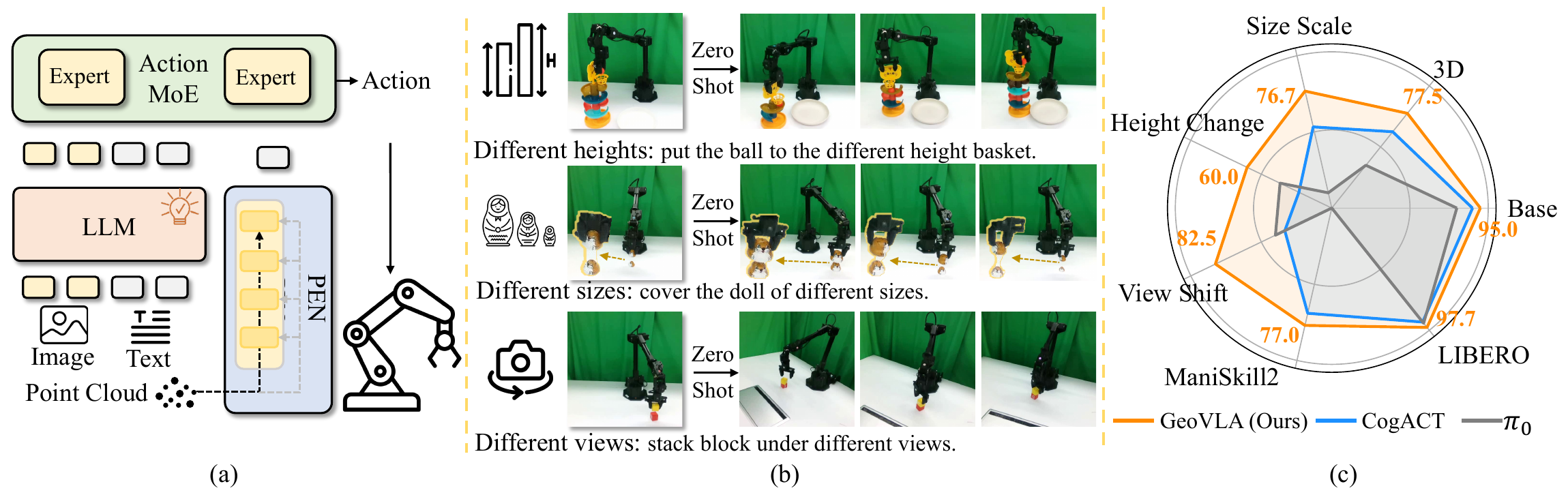}
    \vspace{-10pt}
    \caption{\modelname\ adopts two parallel architecture shown in (a), which additionally extracts 3D geometric information from the point cloud to guide action generation. In this way, \modelname\ shows robust adaptability to height, size, and view variations in (b), and outperforms other methods in (c).}
    \label{fig:01intro}
\end{figure}

In this paper, we present \modelname, a novel VLA framework that integrates 3D information in an elegant way. As shown in Fig.~\ref{fig:01intro}(a), \modelname~incorporates a customized point encoder, \ModulePointEncoder~(\ModulePointEncoderAbbr), and a spatial-aware action expert, \ModuleActionExpert~(\ModuleActionExpertAbbr), to bridge the gap between 2D and 3D modalities. Specifically, \modelname~employs a VLM to process images and language instructions, extracting fused vision-language embeddings. Concurrently, it transforms the depth map into a point cloud and utilizes the \ModulePointEncoderAbbr\ to independently generate 3D geometric embeddings. This dual-path design preserves the pre-trained knowledge and general understanding capabilities of the VLM. Subsequently, these embeddings are concatenated and processed by our novel \ModuleActionExpertAbbr\ module. The \ModuleActionExpertAbbr~employs two specialized experts to jointly model information from both visual and point cloud modalities, ultimately generating precise action sequences. 
In this way, our \modelname\ achieves the multi-modal alignment and superior performance.

We evaluate \modelname\ through both simulation and real-world experiments. In LIBERO~\citep{liu2023libero}, our \modelname~outperforms previous SoTA method OpenVLA-OFT~\citep{kim2025fine} by 2.4\%. For ManiSkill2 ~\citep{mu2021maniskill}, it outperforms Dita~\citep{hou2025dita} by 11\%. In real-world experiments, \modelname\ achieves an average success rate of 86.3\% in 8 tasks, outperforming $\pi_0$~\citep{black2024pi_0} by 28.8\%. Furthermore, \modelname\ exhibits superior robustness in scenarios requiring height adaptation, scale awareness and viewpoint generalization as shown in Fig.~\ref{fig:01intro}(b). 

In summary, our contributions are as follows:

\begin{itemize}
    \item We propose \modelname, a novel Vision-Language-Action (VLA) framework that incorporates both visual and point cloud modalities. Unlike prior work focusing solely on 2D features, \modelname\ explicitly processes multi-modal inputs through parallel branches. This design equips the model with stronger spatial understanding and geometric awareness, maintaining generalization capabilities on height adaptability, scale awareness and viewpoint invariance.

    \item We introduce the \ModulePointEncoder~(\ModulePointEncoderAbbr) and \ModuleActionExpert~(\ModuleActionExpertAbbr). \ModulePointEncoderAbbr\ extracts discriminative geometry-aware features, while the \ModuleActionExpertAbbr\ module leverages modality-specific experts to integrate visual and geometric cues effectively.

    \item Our proposed \modelname\ achieves state-of-the-art performance on the LIBERO and ManiSkill2 benchmarks. It demonstrates significant advantages in incorporating 3D perception, exhibiting robustness in real-world robotic tasks, as shown in Fig.~\ref{fig:01intro}(c).
\end{itemize}

\section{Related Works}
\textbf{Vision Language Action (VLA) Models:} VLA aims to translate natural language instructions and visual observations into executable robot actions, typically building on pretrained vision–language models (VLMs). Early approaches such as RT-1~\citep{brohan2022rt} and RT-2~\citep{brohan2023rt} adopt a language-modeling perspective, discretizing actions into tokens and employing autoregressive decoders for token generation. OpenVLA~\citep{kim2024openvla} extends this paradigm by introducing a large-scale multi-task instruction-following corpus.  
Recent works such as $\pi_0$ \citep{black2024pi_0} Rdt-1b \citep{liu2024rdt} and CogACT \citep{li2024cogact} improve the model by generating continuous actions through flow matching~\citep{lipman2022flow} or diffusion~\citep{ho2020denoising}. By decoupling action representation from tokenization, they achieve more accurate modeling of motion trajectories. However, most VLA models rely solely on 2D RGB inputs, limiting their capacity to reason about spatial structure and depth, which are crucial in real-world scenarios.

\textbf{3D Perception in VLA Models:} With the advancement of 3D foundation models~\citep{wang2024embodiedscan, zhu2024llava, zheng2025densegrounding, zhang2025grounding}, recent studies~\citep{shridhar2023perceiver,jia2024lift3d,goyal2024rvt,ze2024dp3} have explored various 3D representations to enhance robotic manipulation. Meanwhile, the emergence of VLA models, which leverage the pretraining power of VLMs, has opened new directions for generalizable policy learning. But how to integrate 3D perception into VLA models is still a  growing research focus, driven by the limitations of purely 2D inputs in physical environments. Several works \citep{zhen20243d, qu2025spatialvla, Bhat2025cavla, li2025pointvla} aim to make VLA models geometry-aware by introducing 3D features such as depth maps, point clouds, or spatial position embeddings. For example, 3D-VLA~\citep{Zhen20243DVLA} and 3D‑CAVLA~\citep{Bhat2025cavla} directly encode the 3D features as the embeddings to the VLM. While SpatialVLA~\citep{qu2025spatialvla} introduces spatial position embeddings derived from point clouds to equip the visual embeddings. These methods improve spatial reasoning capabilities, but often disrupt the alignment between vision representations and VLMs. Recovering this alignment typically demands additional embodied instruction tuning stages, which can be costly or laborious. For instance, ACT~\citep{dong2022autoencoders} mitigates the need for extensive task-specific fine-tuning on diverse multitask data to strong cross-task generalization. Similarly, LLARVA~\citep{niu2024llarva} highlights the difficulty of incorporating voxel or point cloud representations into existing vision‑language structures and resorts to abstract 2D visual traces to maintain compatibility. These adaptations underscore the trade-off between injecting spatial knowledge and preserving pretrained alignment during deployment.

\textbf{Modality-Aware Action Experts and Point Cloud Fusion:} To preserve the alignment of VLA models while incorporating 3D information, a complementary line of work explores injecting geometry into action heads rather than modifying visual backbones. PointVLA~\citep{li2025pointvla} introduces a zero-initialized 3D feature injector that injects point cloud features into export policy. This design avoids the need to retrain action experts. However, it hinders adaptation to the newly introduced point cloud modality.

Our proposed method, \modelname, distinguishes itself by employing a specialized point encoder (\ModulePointEncoder) to process 3D inputs independently, and a spatial-aware action expert (\ModuleActionExpert) that explicitly models the interaction between visual-language and point cloud representations through modality-specific experts. This design preserves pretrained knowledge, enables more expressive 3D fusion, and improves performance in both simulation benchmarks and real-world robotic tasks.

\section{Methodology}\label{sec:method}

\subsection{Problem Definition}

In the general formulation of vision-language-action (VLA) models, the input consists of two modalities: a visual observation and a natural language instruction. The output is a sequence of actions. Specifically, the visual observation $V$ is typically an RGB image captured from a fixed or egocentric camera, while the language instruction $L$ describes the task to be performed. Given $V$ and $L$, the policy $p$, represented by a pre-trained VLA model, generates a sequence of actions $a_{1:T}$, where $T$ is a hyperparameter that defines the size of the action chunk. Formally,

\begin{equation}\label{eq1}
\begin{aligned}
a_{1:T} \sim p(a_{1:T} \mid V, L)
\end{aligned}
\end{equation}

Each action $a_t$ in the sequence, where $1 \leq t \leq T$, represents a relative movement from the current state of robot, and is typically parameterized as follows:

\begin{equation}\label{eq2}
\begin{aligned}
a_{t} = (\Delta x, \Delta y, \Delta z, \Delta \alpha, \Delta \beta, \Delta \gamma, g)
\end{aligned}
\end{equation}

Here, $(\Delta x, \Delta y, \Delta z)$ denotes the relative translation, $(\Delta \alpha, \Delta \beta, \Delta \gamma)$ denotes the relative rotation (\textit{e.g.}, in Euler angles), and $g$ is the absolute gripper command.

To enhance the spatial perception capabilities of VLA models, we incorporate an additional point cloud $P$, typically obtained from RGB-D camera. The policy then conditions on the visual input $V$, the point cloud $P$, and the language instruction $L$ to generate the action sequence:

\begin{equation}\label{eq4}
\begin{aligned}
a_{1:T} \sim p(a_{1:T} \mid V, P, L)
\end{aligned}
\end{equation}

\subsection{Overview}
\label{sec:pipeline}

\begin{figure}[t]
    \centering
    \includegraphics[width=\textwidth]{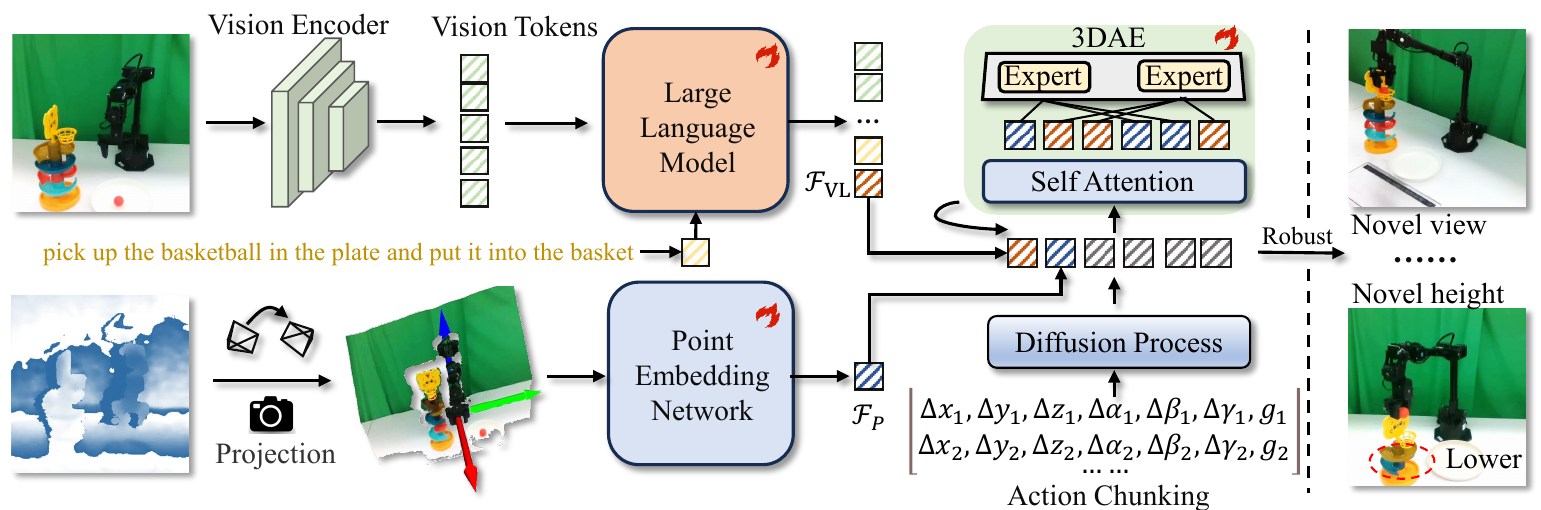}
    \caption{\textbf{Overview of \modelname.} RGB images with language instructions are processed by a VLM to produce vision–language features $\mathcal{F}_{VL}$, while depth maps are reprojected into point clouds and encoded by \ModulePointEncoderAbbr\ as geometric features $\mathcal{F}_{P}$. Both modalities are combined in \ModuleActionExpertAbbr\ to progressively generate robot actions.}
    \label{fig:architecture}
\end{figure}

As illustrated in Fig.~\ref{fig:architecture}, \modelname~is an end-to-end VLA framework that processes visual and geometric information simultaneously. The pipeline comprises three key components: a VLM for general understanding of visual and language modalities, a \ModulePointEncoder~(\ModulePointEncoderAbbr) for extracting fine-grained 3D geometric features, and a \ModuleActionExpert~(\ModuleActionExpertAbbr) for generating action chunk sequences. 

Specifically, given the image $V$ and the language instruction $L$, \modelname~leverages a pre-trained 2D VLM (\textit{e.g.}, Primatic~\citep{karamcheti2024prismatic} to extract a general understanding $\mathcal{F}_{VL}$ of current environment. Concurrently, the depth map $D$ is projected into a 3D point cloud $P$ using the camera parameters. The point cloud is processed by the \ModulePointEncoderAbbr, which is designed to extract a 3D feature $\mathcal{F}_{P}$ centered on the end-effector, capturing critical structural and spatial cues.
The extracted features $\mathcal{F}_{VL}$ and $\mathcal{F}_{P}$ are then concatenated and serve as conditional inputs to the \ModuleActionExpertAbbr. Based on denoising diffusion principles, \ModuleActionExpertAbbr\ starts from a noisy action sequence and progressively refines the noise, guided by the multi-modal context. The mixed expert architecture enables specialized processing of the heterogeneous features $\mathcal{F}_{VL}$ and $\mathcal{F}_{P}$. Each expert learns to capture complementary aspects, thereby enhancing the capacity to generate context-aware and precise action sequences.

\subsection{\ModulePointEncoder}
\label{sec:pcd}

\begin{figure*}[h]
\centering
\includegraphics[width=1.0\linewidth]{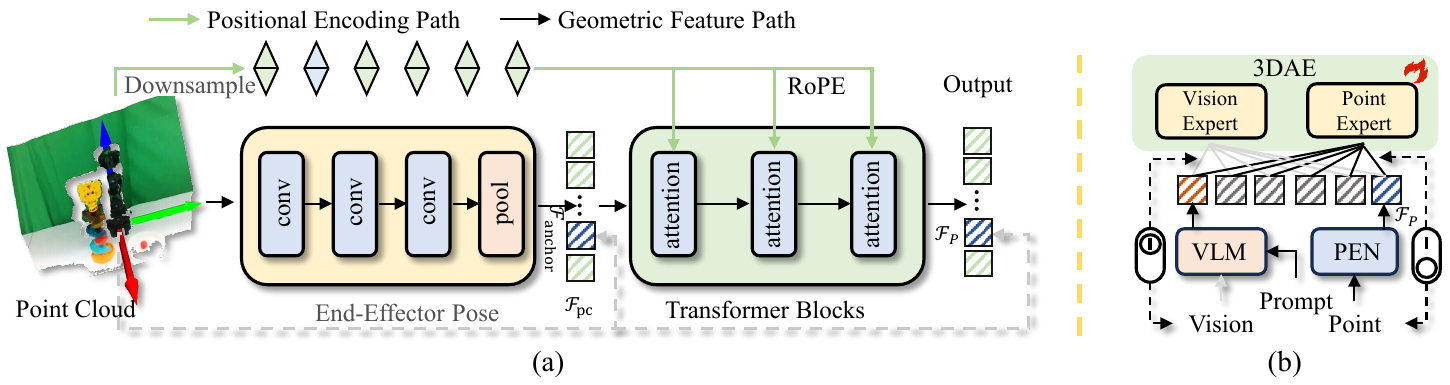}
\caption{\textbf{Dual-path \ModulePointEncoder}. In (a) \ModulePointEncoder processes the point cloud through two parallel paths: geometric feature path using large-kernel convolutions, and a positional encoding path leveraging RoPE to preserve 3D spatial information. In (b) only the selected $\mathcal{F}_{P}$ is send to the action export along with visual feature.}
\label{fig:pen}
\end{figure*}

Due to the inherent noise in the original depth map, extracting compact and clean geometric features is critically important. We propose \ModulePointEncoder\ (\ModulePointEncoderAbbr), a novel geometric point encoder designed for robotic manipulation that captures fine-grained 3D structural cues centered at the end-effector. 

As illustrated in Fig.~\ref{fig:pen}(a), \ModulePointEncoderAbbr\ first converts the raw depth map captured by an RGB-D camera (\textit{e.g.}, RealSense) into a point cloud $P$ expressed in the end-effector coordinate system, where the current end-effector position of robot is at the origin. The encoder then adopts a dual-path architecture:
\textit{(1) Geometric feature path.} A lightweight CNN equipped with multi-layer large-kernel convolutions and local pooling encodes the point cloud into patch-level geometric tokens: $\mathcal{F}_\mathrm{pc} \in \mathbb{R}^{N \times C}$, where $N$ is the number of tokens and $C$ is the feature dimension, followed by the transformer blocks to aggregate the useful global information. \textit{(2) Positional encoding path.} The original point cloud is downsampled to match the size of $\mathcal{F}_\mathrm{pc}$ and guides the position information by rotary positional encoding (RoPE)~\citep{su2024roformer}. 

Subsequently, \ModulePointEncoderAbbr\ selects the token corresponding to the coordinate origin (\textit{e.g., the end-effector token}) as the anchor token $\mathcal{F}_\mathrm{anchor}$ (colored blue in Fig.~\ref{fig:pen}(a)) and feeds the geometric tokens $\mathcal{F}_\mathrm{pc}$ into transformer blocks, where the tokens interact under the guidance of manually generated RoPE. Finally, we only select the updated anchor token from the last transformer layer as the encoded feature $\mathcal{F}_{P}$ to guide the action export as shown in Fig.~\ref{fig:pen}(b). 

This spatial anchor design offers two key benefits: \textit{(1) Focused representation learning.}  The attention mechanism concentrates on manipulation-relevant regions, improving feature extraction efficiency. \textit{(2) Explicit spatial relationship modeling.} Critical geometric relationships between the end-effector and surrounding objects are captured, enabling precise contact prediction.
By delivering localized yet context-aware 3D structural features to downstream action policies, \ModulePointEncoderAbbr\ facilitates spatially grounded decision-making with improved physical coherence.

\subsection{\ModuleActionExpert}
\label{sec:moe}
As illustrated in Fig.~\ref{fig:architecture}, the \ModuleActionExpert~(\ModuleActionExpertAbbr) employs the Diffusion Transformers (DiTs)~\citep{peebles2023scalable} architecture designed to process concatenated multi-modal tokens ($\mathcal{F}_{VL}$ and $\mathcal{F}_{P}$) and generate action sequences. During training, the future action sequence $(\Delta x, \Delta y, \Delta z, \Delta \alpha, \Delta \beta, \Delta \gamma, g)$, sampled from recorded observations, is gradually perturbed to generate a noisy sequence by diffusion process. The noisy actions are served as action tokens and concatenated with the multi-modal tokens. The combined tokens are fed into the diffusion transformer for interaction, ultimately predicting the added noise. During inference, we first sample a noise distribution using DDIM~\citep{song2020denoising} and then, conditioned on multi-modal tokens, progressively recover the desired action sequence.

However, effectively integrating features from different modalities presents a significant challenge. To leverage the complementary strengths of vision-language features $\mathcal{F}_{VL}$ and geometric features $\mathcal{F}_{P}$, we introduce a novel Mixture-of-Experts (MoE) architecture within the feed-forward networks (FFNs) of our diffusion transformer action head \ModuleActionExpertAbbr. This design enables specialized processing for each modality, allowing the model to more effectively utilize multi-modal conditional information.

Since the VLM branch is pre-trained while the point cloud branch is initialized from scratch, our experimental results indicate that directly applying a dynamic routing mechanism inherently biases the model toward the VLM branch. To mitigate this imbalance, we introduce a static routing strategy. During training, we randomly drop one modality in each iteration, resulting in three distinct input configurations: (1) vision–language features only, (2) language and geometric features, where RGB image tokens are removed before being fed into the VLM, and (3) the full multi-modal input. In \ModuleActionExpertAbbr, the activation of each expert is deterministically governed by the presence and relevance of input modalities. 
This static, purpose-driven routing strategy within the MoE architecture ensures that the unique strengths of each modality are effectively leveraged. At the same time, it maintains the powerful generative capabilities of the DiT, enabling robust and comprehensive action chunk generation for robotic control. 

\section{Simulation Experiments}
In this section, we conduct a thorough evaluation of \modelname~ on two widely-used robotic manipulation benchmarks, LIBERO~\citep{liu2023libero} and ManiSkill2~\citep{mu2021maniskill}, to assess the effectiveness and scalability of our approach across diverse tasks.

\noindent\textbf{LIBERO} consists of five task suites, spanning four unique scenes. Each suite is designed to evaluate specific capabilities:
\begin{itemize}
\item \textbf{LIBERO-Spatial} focuses on placing the same object in different positions.
\item \textbf{LIBERO-Object} involves placing different objects into a box within a fixed scene layout.
\item \textbf{LIBERO-Goal} evaluates the ability to perform diverse operations in a fixed layout.
\item \textbf{LIBERO-Long}, also called LIBERO-10, which targets 10 long-horizon tasks involving varied scenes and operations.
\item \textbf{LIBERO-90} is an expanded version of LIBERO-10, presents a more challenging benchmark.
\end{itemize}
The more visualization of LIBERO benchmark can be found in Fig~.\ref{fig:07benchmark}(a).

\noindent\textbf{ManiSkill2} focuses more on basic pick-and-place capabilities. Following Dita~\citep{hou2025dita}, we evaluate our approach on five representative tasks: \textit{PickCube}, \textit{StackCube}, \textit{PickSingleYCB}, \textit{PickSingleEGAD}, and \textit{PickClutterYCB}.
Except for \textit{StackCube}, each task requires the robot to grasp a specified object and place it at a designated 3D position indicated by a green marker, making them well-suited to evaluate 3D perception and spatial reasoning.
Furthermore, \textit{PickSingleYCB} and \textit{PickSingleEGAD} introduce greater challenges by varying the objects, with a combined total of around 1700 distinct objects. \textit{PickClutterYCB} is even more challenging, requiring the robot to identify and select the correct object from a cluttered scene containing up to 74 different types of YCB~\citep{xiang2017posecnn} objects.
The more visualizations of ManiSkill2 benchmark can be found in Fig~.\ref{fig:07benchmark}(b).

\subsection{Implementation Details}

The VLM component in \modelname\ is Prismatic-7B~\citep{karamcheti2024prismatic}, initialized using the pre-trained weights released by OpenVLA~\citep{kim2024openvla}, which is trained on the large-scale Open X-Embodiment dataset~\citep{o2024open}. In contrast, both the proposed \ModulePointEncoderAbbr\ and \ModuleActionExpertAbbr\ are randomly initialized.
All experiments are conducted on 8 NVIDIA A100 GPUs using the Fully Sharded Data Parallel (FSDP) strategy. Each GPU processes a batch size of 32, resulting in a total batch size of 256. Optimization is performed using the AdamW optimizer \citep{loshchilov2019adamw} with a constant learning rate of $2 \times 10^{-5}$. Mixed-precision training \citep{micikevicius2018mixed} is enabled to improve efficiency and reduce memory consumption.
The dataset is loaded via TensorFlow Datasets (TFDS), with a shuffle buffer size of 10,000. No data augmentation is applied during training. For action modeling, we adopt a chunking strategy with a fixed length of $T=16$.

For the LIBERO benchmark~\citep{liu2023libero}, \modelname\ processes only a single main camera view together with the point cloud map derived from the corresponding depth image, without incorporating any additional inputs (\textit{e.g.}, pose states or wrist-view images). In contrast, for ManiSkill2~\citep{mu2021maniskill}, additional information is provided, including the proprioceptive state, the gripper state, and the target marker position since there are some occlusions in the image. Training runs for both benchmarks approximately 20,000 steps (about 20 hours), corresponding to roughly 6 epochs on LIBERO and 2 epochs on ManiSkill2. During inference, we follow the standard evaluation protocol: each LIBERO task is assessed over 50 independent episodes, while each ManiSkill2 task is evaluated over 20 episodes. All results are reported in terms of success rate (SR).

\subsection{Main Results}
\noindent\textbf{LIBERO.}
Tab.~\ref{tab:libero} compares \modelname\ with several strong baselines on the LIBERO~\citep{liu2023libero} benchmark. \modelname\ delivers the best performance across all tasks. In particular, for the LIBERO-Long and LIBERO-90 tasks, \modelname\ achieves success rates of 96.6\% and 97.7\%, improving by 5.9\% and 5.6\%, respectively. Overall, \modelname\ achieves an average success rate of 97.7\%, outperforming both CogACT~\citep{li2024cogact} (93.2\%) and OpenVLA-OFT~\citep{kim2025fine} (95.3\%).

\begin{table}[t]
    \centering
    \setlength{\tabcolsep}{10pt} 
    \caption{\textbf{Results on LIBERO.} * denotes methods that incorporate extra proprioceptive states or wrist-camera images. Results of CogACT are reproduced by us. For methods not evaluated on LIBERO-90, the average success rate across the first four suites is reported.}
    \resizebox{\textwidth}{!}{
    \begin{tabular}{l|ccccc|c}
        \toprule
        Method & Spatial & Object & Goal & Long & LIBERO-90 & Avg. \\
        \midrule
        Octo~\citep{team2024octo} & 78.9 & 85.7 & 84.6 & 51.1 & - & 75.1 \\ 
        OpenVLA~\citep{kim2024openvla} & 84.7 & 88.4 & 79.2 & 53.7 & 73.5 & 75.9 \\ 
        SpatialVLA~\citep{qu2025spatialvla} & 88.2 & 89.9 & 78.6 & 55.5 & 46.2 & 71.7 \\ 
        $\pi_0$-FAST*~\citep{pertsch2025fast} & 96.4 & 96.8 & 88.6 & 60.2 & 83.1 & 85.0\\ 
        $\pi_0$*~\citep{black2024pi_0} & 96.8 & \underline{98.8} & 95.8 & 85.2 & - & 94.2\\
        % CogACT~\citep{li2024cogact} & \underline{97.2}  & 98.0 & 93.2 & 90.6 & \underline{95.4} & 94.9 \\
        CogACT~\citep{li2024cogact} & \underline{97.2}  & 98.0 & 90.2 & 88.8 & \underline{92.1} & 93.2 \\
        OpenVLA-OFT*~\citep{kim2025fine} & 96.2 & 98.3 & \underline{96.2} & \underline{90.7} & - & \underline{95.3} \\ 
        \midrule
        \rowcolor{lightgray} \modelname~(Ours) & \textbf{98.4} & \textbf{99.0} & \textbf{96.6} & \textbf{96.6} & \textbf{97.7} & \textbf{97.7} \\
        \bottomrule
    \end{tabular}}
    \label{tab:libero}
    % \vspace{-20pt}
\end{table}

\noindent\textbf{ManiSkill2.} Tab.~\ref{tab:ManiSkill2} presents our results on the ManiSkill2~\citep{mu2021maniskill} benchmark. \modelname\ achieves the highest overall success rate of 77\%, clearly surpassing both CogACT (69\%) and Dita (66\%). Specifically, on basic tasks such as \textit{PickCube}, \modelname\ performs on par with CogACT, the best-performing method. However, on more challenging tasks like \textit{PickClutterYCB}, \modelname\ achieves the best performance, with a success rate of 45\% compared to 36\% for Dita. Other methods struggle to improve their performance as the diversity of objects significantly increases the difficulty of 6D pose estimation. This challenge arises from the increased complexity of spatial position estimation due to object diversity. In contrast, \modelname\ leverages point cloud observations to maintain precise spatial awareness, resulting in substantially higher success rates on these more difficult tasks.

\begin{table}[t]
    \centering
    \vspace{-10pt}
    \caption{\textbf{Results on ManiSkill2.} * denotes our reproduced results. All the results are reported in terms of success rate (\%).}
    \resizebox{\textwidth}{!}{
    \setlength{\tabcolsep}{2pt} 
    \begin{tabular}{l|ccccc|c}
        \toprule
        Method & PickCube & StackCube & PickSingleYCB & PickSingleEGAD & PickClutterYCB & Avg. \\
        \midrule
        OpenVLA*~\citep{kim2024openvla} & 65 & 55 & 0 & 15 & 0 & 27 \\ 
        Dita~\citep{hou2025dita} & 79 & 80 & 62 & 72 & \underline{36} & 66 \\ 
        CogACT*~\citep{li2024cogact} & \textbf{95} & \underline{90} & \underline{65} & \underline{75} & 25 & \underline{69}\\
        % $\pi_0$*~\citep{black2024pi_0} & - & - & - & - & - & -\\
        \midrule
        \rowcolor{lightgray} GeoVLA (Ours) & \underline{90} & \textbf{90} & \textbf{75} & \textbf{85} & \textbf{45} & \textbf{77} \\
        \bottomrule
    \end{tabular}}
    \label{tab:ManiSkill2}
\end{table}

\subsection{Ablation Study}

To assess the impact of key components in our proposed \modelname, we perform a comprehensive ablation study on the LIBERO benchmark, the results of which are summarized in Tab.~\ref{tab:ablation_study}.

\begin{table}[h]
\centering
\vspace{-5pt}
\caption{\textbf{Ablation Study.} For \ModulePointEncoderAbbr, the ablation study demonstrates its efficiency by (a) comparing with other point encoders, (b) the anchor token selection strategy, and (c) analyzing the impact of position embeddings. For \ModuleActionExpertAbbr, the effectiveness of the MoE-based design is evaluated.}
\label{tab:ablation_study}
\setlength{\tabcolsep}{12pt} 
\resizebox{\textwidth}{!}{
\begin{tabular}{l|c|ccccc|c} 
\toprule
\multicolumn{2}{l|}{Method} & Spatial & Object & Goal & Long & LIBERO-90 & Avg. \\ 
\midrule
% \multicolumn{2}{c|}{Baseline} & 97.2\% & 98.0\% & 93.2\% & 90.6\% & 95.4\% & 94.9\% \\
% \midrule
\multirow{3}{*}{(a)} & MLP & 97.6 & 99.0 & 97.8 & 88.4 & 96.1 & 95.8 \\
 & PointNet & 94.8 & 96.2 & 95.4 & 94.8 & 95.0 & 95.2 \\ 
 & \cellcolor{lightgray}\ModulePointEncoderAbbr & \cellcolor{lightgray}98.4 & \cellcolor{lightgray}99.0 & \cellcolor{lightgray}96.6 & \cellcolor{lightgray}96.6 & \cellcolor{lightgray}97.7 & \cellcolor{lightgray}97.7 \\
\midrule
\multirow{3}{*}{(b)} & Max   & 98.2 & 98.2 & 95.4 & 94.6 & 95.0 & 96.3 \\ 
 & Mean & 98.0 & 96.6 & 95.2 & 92.6 & 97.2 & 95.9 \\
 & \cellcolor{lightgray}End Effector & \cellcolor{lightgray}98.4 & \cellcolor{lightgray}99.0 & \cellcolor{lightgray}96.6 & \cellcolor{lightgray}96.6 & \cellcolor{lightgray}97.7 & \cellcolor{lightgray}97.7 \\
\midrule
\multirow{2}{*}{(c)} & 1D PE & 95.2 & 98.2 & 96.4 & 92.0 & 95.2 & 95.4 \\
& \cellcolor{lightgray}RoPE & \cellcolor{lightgray}98.4 & \cellcolor{lightgray}99.0 & \cellcolor{lightgray}96.6 & \cellcolor{lightgray}96.6 & \cellcolor{lightgray}97.7 & \cellcolor{lightgray}97.7 \\
\midrule
\multirow{3}{*}{(d)} & None MoE  & 98.0 & 98.4 & 94.6 & 93.2 & 95.7 & 96.0 \\
 & Dynamic Routing  & 98.4 & 98.4 & 98.4 & 94.8 & 96.6 & 97.3 \\
 & \cellcolor{lightgray}Static Routing& \cellcolor{lightgray}98.4 & \cellcolor{lightgray}99.0 & \cellcolor{lightgray}96.6 & \cellcolor{lightgray}96.6 & \cellcolor{lightgray}97.7 & \cellcolor{lightgray}97.7 \\

\bottomrule
\end{tabular}}
\end{table}

\textbf{Comparison with Other Point Cloud Encoders.} The effect of different point cloud encoders is examined by comparing a 3-layer MLP (\textit{e.g.}, DP3 \citep{ze2024dp3}), PointNet~\citep{qi2017pointnet}, and the proposed \ModulePointEncoderAbbr\ encoder. Tab.~\ref{tab:ablation_study}(a) illustrates that the MLP and PointNet encoders achieve success rates of 95.8\% and 95.2\%, respectively, while the \ModulePointEncoderAbbr\ encoder outperforms them with a success rate of 97.7\%, demonstrating its superior ability to capture geometric structure.

\textbf{Different Anchor Token Selection Strategies.} Three anchor token selection strategies are evaluated in Fig.~\ref{tab:ablation_study}(b): max pooling (\textit{Max}), average pooling (\textit{Mean}), and directly using the end-effector token (\textit{End Effector}) as the pooled representation. The \textit{End Effector} strategy achieves the best results (97.7\%) compared to the \textit{Max} (96.3\%) and \textit{Mean} (95.9\%), suggesting that the end-effector token carries more task-relevant and informative features.

\textbf{Impact of Position Embeddings.} The role of Rotary Positional Embedding (\textit{RoPE}) in the \ModulePointEncoderAbbr\ encoder is examined in Fig.~\ref{tab:ablation_study}(c). Compared to the original 1D learnable position embeddings, incorporating \textit{RoPE} consistently improves performance, increasing the success rate from 95.4\% to 97.7\%, confirming its effectiveness in enhancing spatial encoding.

\textbf{The Effectiveness of MoE Design in \ModuleActionExpertAbbr.} Fig.~\ref{tab:ablation_study}(d) compares the performance of our MoE-based \ModuleActionExpertAbbr\ with non-MoE action heads commonly used in VLA models. Our MoE-based design achieves a performance gain of 96.0\% to 97.7\%. Although dynamic routing is the default in most MoE architectures, our experiments show that static routing performs better, likely due to its reduced bias towards the vision-language modalities during training.

\section{Real-world Experiments}
In this section, we validate \modelname~through real-world robotic experiments involving various 3D manipulation tasks. These experiments demonstrate the robustness and generalization of \modelname, particularly under common deployment variations of the camera viewpoint, height, and object size.

\textbf{Environment Setup.} In our real-world experiments, we employ a WidowX-250s robotic arm with six degrees of freedom and a RealSense-435i depth camera placed approximately 0.8 meters away to capture a third-person viewpoint as shown in Fig~.\ref{fig:real_setup}. The training setup mirrors the simulation experiments, with the training duration shortened to about 8 hours due to the smaller dataset. Each task is evaluated over 10 independent trials during inference.

\subsection{Task Definition}
\label{task_def}
A diverse set of tasks is used to evaluate \modelname, as shown in Fig.~\ref{fig:real_task} of the supplementary material. These tasks include basic manipulation tasks such as \textit{Pick Carrot} (P-Carrot), \textit{Stack Block} (S-Block), \textit{Stack Cup} (S-Cup), and \textit{Insert Circle} (I-Circle), as well as 3D-aware tasks like \textit{Hang Cup} (H-Cup), \textit{Put Basketball} (P-Basketball), \textit{Cover Matryoshka} (C-Matryoshka), and \textit{Put Hairclip} (P-Hairclip), which test spatial perception and precise manipulation.  We also introduce challenging task variations to assess spatial robustness during inference:  
(1) Varying the basket height in \textit{Put Basketball} to test adaptability to target position changes.  
(2) Changing the camera viewpoint in \textit{Stack Block} to evaluate viewpoint invariance.  
(3) Scaling the doll size in \textit{Cover Matryoshka} to assess relative size perception.  
(4) Removing the sponge mat in \textit{Pick Carrot} to test robustness to height variations in carrot placement.
More details are available in the supplementary material.

\subsection{Main Results}

\begin{table}[t]
\centering
\caption{\textbf{Succuss rate (\%) on in-domain tasks.} \modelname\ achieves better performance compared with OpenVLA, $\pi_0$, and CogACT.}
\label{tab:basic_tasks_performance}
    \resizebox{\textwidth}{!}{
    \setlength{\tabcolsep}{1pt} 
    \begin{tabular}{l|cccc|cccc|c}
    \toprule
    Method & P-Carrot & S-Block & S-Cup & I-Circle & H-Cup & P-Basketball & C-Matryoshka & P-Hairclip & Avg. \\
    \midrule
OpenVLA~\citep{kim2024openvla} & 50 & 10 & 40 & 30 & 0 & 20 & 10 & 0 & 20.0 \\
    % SpatialVLA      &                      &                     &                    &                        &                  \\
    $\pi_0$~\citep{black2024pi_0} & \underline{100} & 50 & 90 & \underline{80} &  \underline{50} & 40 & 50 & 0 & 57.5\\
    CogACT~\citep{li2024cogact}  & \underline{100} & \underline{80} & \underline{100} & \underline{80} & \underline{50} & \underline{60} & \underline{70} & \underline{70} & \underline{76.3} \\
    \midrule
    \rowcolor{lightgray} \modelname~(Ours) & \textbf{100} & \textbf{80} & \textbf{100} & \textbf{100} &  \textbf{70} & \textbf{90} & \textbf{70} & \textbf{80} & \textbf{86.3}\\
    \bottomrule
    \end{tabular}}
\end{table}

Evaluation of \modelname\ is conducted on a set of in-domain tasks, including both basic manipulation and 3D-aware tasks, with identical training and inference setups. We also compare our results against several state-of-the-art baselines, including OpenVLA \citep{kim2024openvla}, $\pi_0$ \citep{black2024pi_0} (without pose states and other images as input), and CogACT \citep{li2024cogact}. As shown in Tab.~\ref{tab:basic_tasks_performance}, most models perform well on simple pick-and-place tasks, such as \textit{Pick Carrot}, achieving a 100\% success rate. However, performance drops significantly on tasks that require precise 3D spatial understanding, like \textit{Put Basketball} and \textit{Pick Hairclip}. In these tasks, our approach consistently outperforms others, showing better robustness and spatial awareness. Specifically, Our model achieves average success rates of 95.0\% on basic tasks and 77.5\% on 3D-aware tasks, yielding an overall average success rate of 86.3\%, which outperforms $\pi_0$ and CogACT 28.8\% and 10.0\% respectively by a clear margin.

\subsection{Variant Results}

\begin{figure*}[t]
\centering
\includegraphics[width=1.0\linewidth]{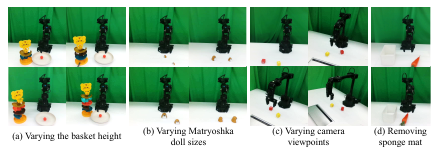}
\caption{\textbf{Task variation visualization.} Four types of variation are conducted: (a) basket height, (b) Matryoshka doll sizes, (c) camera viewpoints, and (d) presence/absence of the sponge mat.}
\label{fig:variation}
\end{figure*}

To further assess the robustness and generalization ability of our \modelname, we introduce controlled task variations that were not seen during training. As shown in Fig.~\ref{fig:variation}, for each variation, a specific task setup is modified while keeping all other conditions unchanged.

\noindent\textbf{\textit{Put Basketball} variation Task.} The training data only includes a basket positioned at the fifth layer (\textit{Base}). During inference, the model is tested with varying basket heights at the third (\textit{L2}), fourth (\textit{L1}), and sixth layers (\textit{H1}). As shown in Tab.~\ref{tab:ball_height}, the 2D-VLA models like CogACT experience a performance drop when the basket height deviates from the training condition, achieving success rates of only 20\% and 30\% at \textit{L2} and \textit{L1}, respectively, while $\pi_0$ achieves a similar success rate compared to \textit{Base}, struggling at picking precisely.  In contrast, \modelname\ exhibits stronger generalization, keeping reasonable performance across all variations and achieving a 60\% success rate even at the highest basket position (\textit{H1}). This shows the robustness of \modelname\ to spatial distribution shifts.

\noindent\textbf{\textit{Cover Matryoshka} variation Task.} The model is trained using a medium-sized doll (\textit{Base}). During inference, we evaluate the model by scaling the doll to different sizes: a smaller size (\textit{S1}), a slightly larger size (\textit{L1}), and a significantly larger size (\textit{L2}), to assess its robustness to scale variations. The results are summarized in Tab.~\ref{tab:doll_size}. Compared to other 2D-VLA models, our \modelname\ consistently achieves higher success rates at larger sizes (\textit{L1} and \textit{L2}).

\begin{table}[t]
\centering 
\begin{minipage}[t]{0.48\textwidth}
    \centering
    \captionof{table}{\textbf{Success rate (\%) under varying basket heights.} 
    L2 and L1 indicate the basket is placed 2 and 1 layers lower than the training (Base), respectively. H1 indicates 1 layer higher.}
    \label{tab:ball_height}
    \setlength{\tabcolsep}{5pt}
    \resizebox{\textwidth}{!}{
    \begin{tabular}{l|cccc}
    \toprule
    Method & L2 & L1 & Base & H1 \\
    \midrule
    $\pi_0$~\citep{black2024pi_0}  & 50        & 40        & 50          & 20        \\
    CogACT~\citep{li2024cogact}  & 20        & 30        & 60          & 20        \\
    \rowcolor{lightgray} \modelname~(Ours) & \textbf{50} & \textbf{70}        & \textbf{90}          & \textbf{60}        \\
    \bottomrule
    \end{tabular}}
\end{minipage}
\hfill 
\begin{minipage}[t]{0.48\textwidth}
    \centering
    \captionof{table}{\textbf{Success rate (\%) under varying doll sizes.} 
    Base refers to the training size. S1 denotes a smaller size, while L1 and L2 represent slightly and significantly larger dolls, respectively.}
    \label{tab:doll_size}
    \setlength{\tabcolsep}{5pt}
    \resizebox{\textwidth}{!}{
    \begin{tabular}{l|cccc}
    \toprule
    Method & S1 & Base & L1 & L2 \\
    \midrule
    $\pi_0$~\citep{black2024pi_0}  & 10        & 50        & 20          & 0        \\
    CogACT~\citep{li2024cogact} & 70        & 70        & 40          & 50        \\
    \rowcolor{lightgray} \modelname~(Ours) & \textbf{70}        & \textbf{70}        & \textbf{80}          & \textbf{80}        \\
    \bottomrule
    \end{tabular}}
\end{minipage}
\end{table}

\noindent\textbf{\textit{Stack Block} variation Task.} we jointly train the models using only the main camera, alongside other tasks with a side camera to enhance generalization. During inference, all the models are evaluated under specific camera views. As shown in Tab.~\ref{tab:block_view}, both CogACT~\citep{li2024cogact} and our \modelname\ perform well when the camera pose matches the training setting (\textit{Base}). However, when the camera shifts to 45°, \modelname\ maintains high performance, while the performance of CogACT~\citep{li2024cogact} drops significantly.

\noindent\textbf{\textit{Pick Carrot} variation Task.} The sponge mat present during training is removed during inference, resulting in a lower placement of the carrot. While most methods attempt to grasp the carrot from above, leading to failure, our geometric-aware \modelname\ consistently achieves more stable and successful grasps, demonstrating stronger generalization to this change.

\begin{table}[t]
\centering
\begin{minipage}[t]{0.51\textwidth}
    \centering
    \captionof{table}{\textbf{Success rate (\%) under varying camera views.} Evaluation was conducted at 15°, 30°, and 45° viewpoints.}
    \label{tab:block_view}
    \setlength{\tabcolsep}{5pt}
    \resizebox{\textwidth}{!}{
    \begin{tabular}{l|cccc}
    \toprule
    Method & Base & 15° & 30° & 45° \\
    \midrule
    $\pi_0$ \citep{black2024pi_0}       & 50        & 50        & 40          & 30        \\
    CogACT \citep{li2024cogact}  & 80        & 60        & 40          & 0        \\
    \rowcolor{lightgray} \modelname~(Ours)   & \textbf{90}        & \textbf{90}        & \textbf{80}          & \textbf{70}        \\
    \bottomrule
    \end{tabular}}
\end{minipage}
\hfill 
\begin{minipage}[t]{0.45\textwidth}
    \centering
    \captionof{table}{\textbf{Success rate (\%) with or without the sponge mat} in the \textit{Pick Carrot} Task. \modelname\ shows strong generalization.}

    \label{tab:carrot_height}
    \setlength{\tabcolsep}{6pt}
    \resizebox{\textwidth}{!}{
    \begin{tabular}{l|cc}
    \toprule
    Method & with & without \\
    \midrule
    $\pi_0$ \citep{black2024pi_0}  & 100        & \textbf{60}        \\
    CogACT \citep{li2024cogact}  & 100        & 10        \\
    \rowcolor{lightgray} \modelname~(Ours)   & \textbf{100}        & 50        \\
    \bottomrule
    \end{tabular}}
\end{minipage}
\end{table}

\section{Conclusion}

In this paper, we investigated how to effectively integrate 3D representations into Vision-Language Action (VLA) models and proposed \modelname. To extract compact representations from noisy point clouds, \modelname\ employs a dual-path \ModulePointEncoder, which selects a token corresponding to the end-effector position as an anchor for geometric features. Coupled with vision–language features from a pre-trained VLM, a diffusion-based action head, \ModuleActionExpert, processes both modalities to progressively generate action chunks. To mitigate the bias toward vision–language inputs, we further introduced multiple experts and a manually designed expert-balancing strategy. We evaluated \modelname\ in both simulation and real-world experiments, where it demonstrates strong performance. Notably, compared with 2D-VLA baselines, \modelname\ achieves significantly better results under 3D-related variations such as camera viewpoint changes, highlighting the effectiveness of integrating 3D representations into VLA models.
{\small\bibliographystyle{ieeenat_fullname}\bibliography{main}}

\clearpage
\setcounter{page}{1}
\appendix

\section{Simulation Environments}
\begin{figure*}[ht]
\centering
\includegraphics[width=1.0\linewidth]{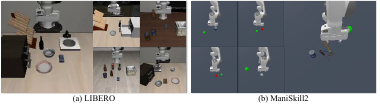}
\vspace{-15pt}
\caption{\textbf{Simulation benchmarks}. The LIBERO benchmark (a) contains various scenes and tasks, and the ManiSkill2 benchmark Pick-and-Place tasks (b) are required to pick an object to the specific location marked by a green point in a 3D space.}
\label{fig:07benchmark}
\end{figure*}

Fig. \ref{fig:07benchmark} gives the overall visualization of the two simulation benchmarks including LIBERO and ManiSkill2 in our experiments. The LIBERO benchmark consists of five task suites focusing on different skills, while the ManiSkill2 benchmarks focus primarily on pick-and-place scenarios.

\section{Visualization of Simulation Results}
\begin{figure*}[ht]
\centering
\includegraphics[width=1.0\linewidth]{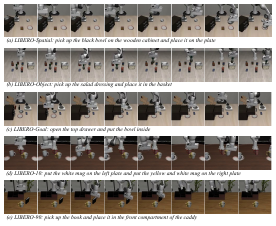}
\caption{\textbf{Qualitative results of GeoVLA} on the five task suites of LIBERO benchmark. }  
\label{fig:07visual_libero}
\end{figure*}

\begin{figure*}[ht]
\centering
\includegraphics[width=1.0\linewidth]{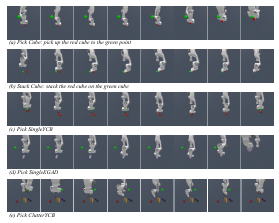}
\caption{\textbf{Qualitative results of GeoVLA} on the five tasks of ManiSkill2 benchmark. }  
\label{fig:07visual_maniskill}
\end{figure*}

Fig.~\ref{fig:07visual_libero} and Fig.~\ref{fig:07visual_maniskill} present qualitative results on representative tasks from LIBERO and ManiSkill2, respectively. These visualizations demonstrate that our approach can accurately recognize and interact with the objects in different tasks and different environments, highlighting its strong performance and adaptability.

\section{Real-World Environments}
\begin{figure*}[t]
\centering
\includegraphics[width=0.6\linewidth]{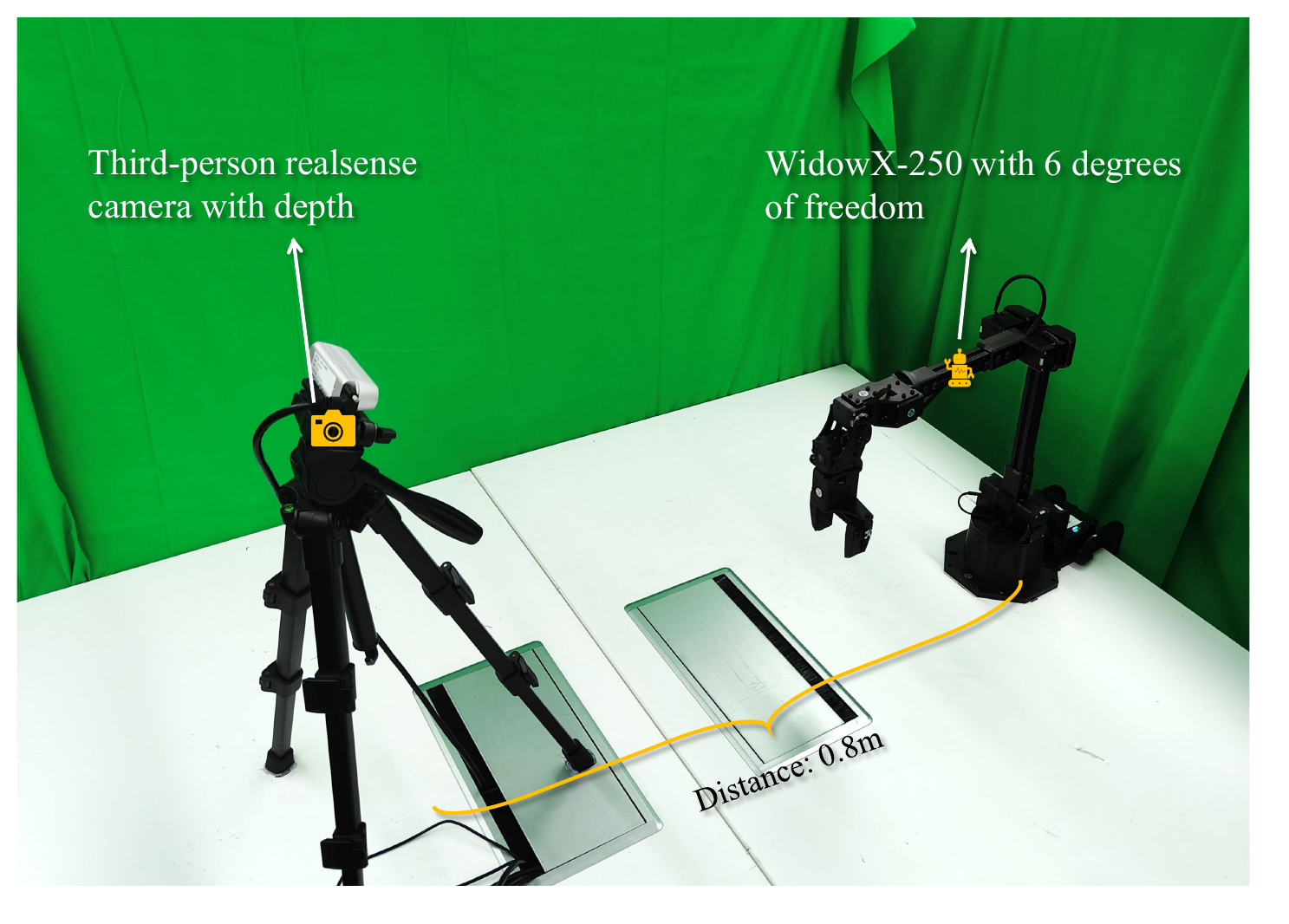}
\caption{\textbf{Setup of real experiments}. We  describe our real-world experimental setup, which includes a depth camera and a robotic arm. }  
\label{fig:real_setup}
\vspace{0pt}
\end{figure*}

\begin{figure*}[t]
\centering
\includegraphics[width=1.0\linewidth]{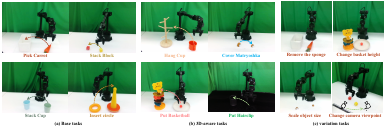}
\caption{\textbf{Tasks in real-world experiments}. We divide the tasks into two categories: (a) base tasks designed to evaluate essential pick-and-place abilities; (b) 3D-aware tasks that require more accurate spatial reasoning and object pose estimation; and (c) task variations used at inference time to assess robustness and generalization.}
\label{fig:real_task}
\end{figure*}

In our real-world experiments, we employ a WidowX-250s robotic arm with six degrees of freedom and a RealSense-435i depth camera placed approximately 0.8 meters away to capture a third-person viewpoint, as shown in Fig. \ref{fig:real_setup}. The basic tasks and 3D-aware tasks we conduct in our real-world experiments are shown below and the overall visualization is shown in Fig. \ref{fig:real_task}.

\noindent\textbf{Base Tasks.}
We consider four fundamental manipulation tasks designed to evaluate basic pick-and-place capabilities:
(1) \textit{Pick Carrot}: The robot is required to grasp a slender carrot and place it into a box;
(2) \textit{Stack Block} and (3) \textit{Stack Cup}: The robot needs to pick up a yellow block or a green cup and stack it onto another object;
(4) \textit{Insert Circle}: The robot must pick up a circular object and insert it into a vertical pillar.
These base tasks are jointly trained to improve training efficiency and facilitate shared skill learning.

\noindent\textbf{3D-Aware Tasks.}
We also design a set of more challenging tasks that require accurate perception of object poses and spatial relationships in 3D space:
(1) \textit{Hang Cup}: The robot must grasp a cup and precisely hang it on a narrow holder;
(2) \textit{Put Basketball}: The robot picks up a basketball and places it into a basket;
(3) \textit{Cover Matryoshka}: The robot needs to pick up a larger Matryoshka doll and cover a smaller one;
(4) \textit{Put Hairclip}: The robot is required to pick up a \textbf{black} hairclip and place it into a bucket located inside a \textbf{black} desk—this presents significant challenges for RGB-based perception due to low color contrast and background ambiguity.
Unlike the base tasks, each 3D-aware task is trained independently to avoid task interference, ensuring that generalization capabilities are learned from the model architecture and spatial understanding.

\section{Visualization of Real-World Results}

\begin{figure*}[th]
\centering
\includegraphics[width=1.0\linewidth]{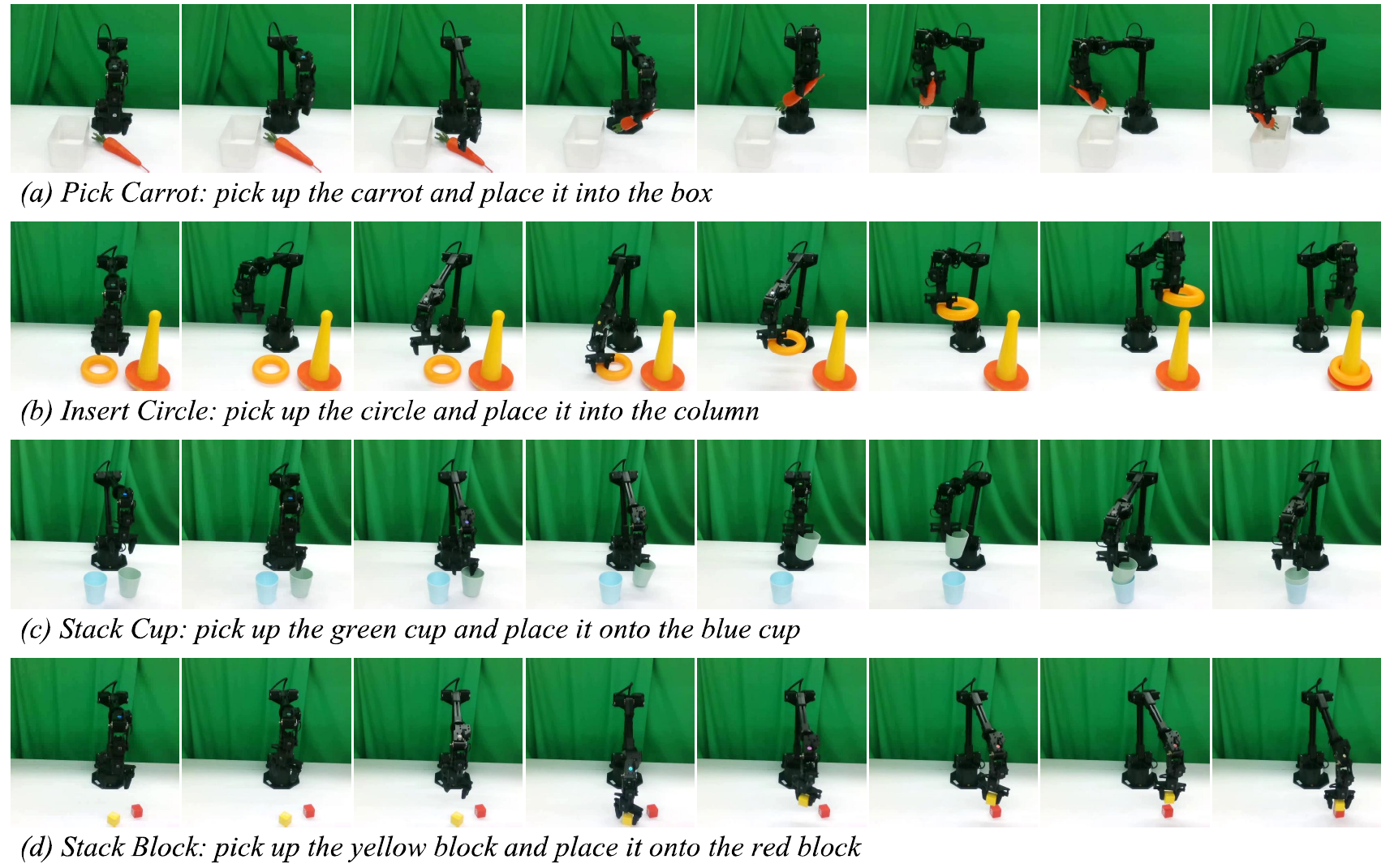}
\caption{\textbf{Examples of base task executions with GeoVLA.}}
\label{fig:07visual_basic}
\end{figure*}

\begin{figure*}[th]
\centering
\includegraphics[width=1.0\linewidth]{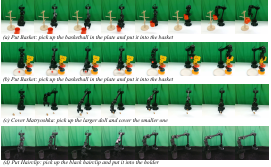}
\caption{\textbf{Examples of 3D-aware task executions with GeoVLA.}}
\label{fig:07visual_3d}
\end{figure*}

Fig.~\ref{fig:07visual_basic} and Fig.~\ref{fig:07visual_3d} illustrate the qualitative performance of GeoVLA on base tasks and 3D-aware tasks, respectively. GeoVLA demonstrates high precision and success rates across all tasks. In particular, for 3D-aware tasks such as the \textit{Put Hairclip} task, GeoVLA effectively leverages point cloud inputs to extract detailed geometric structures, enabling robust performance even in the absence of RGB inputs.

\begin{figure*}[th]
\centering
\includegraphics[width=1.0\linewidth]{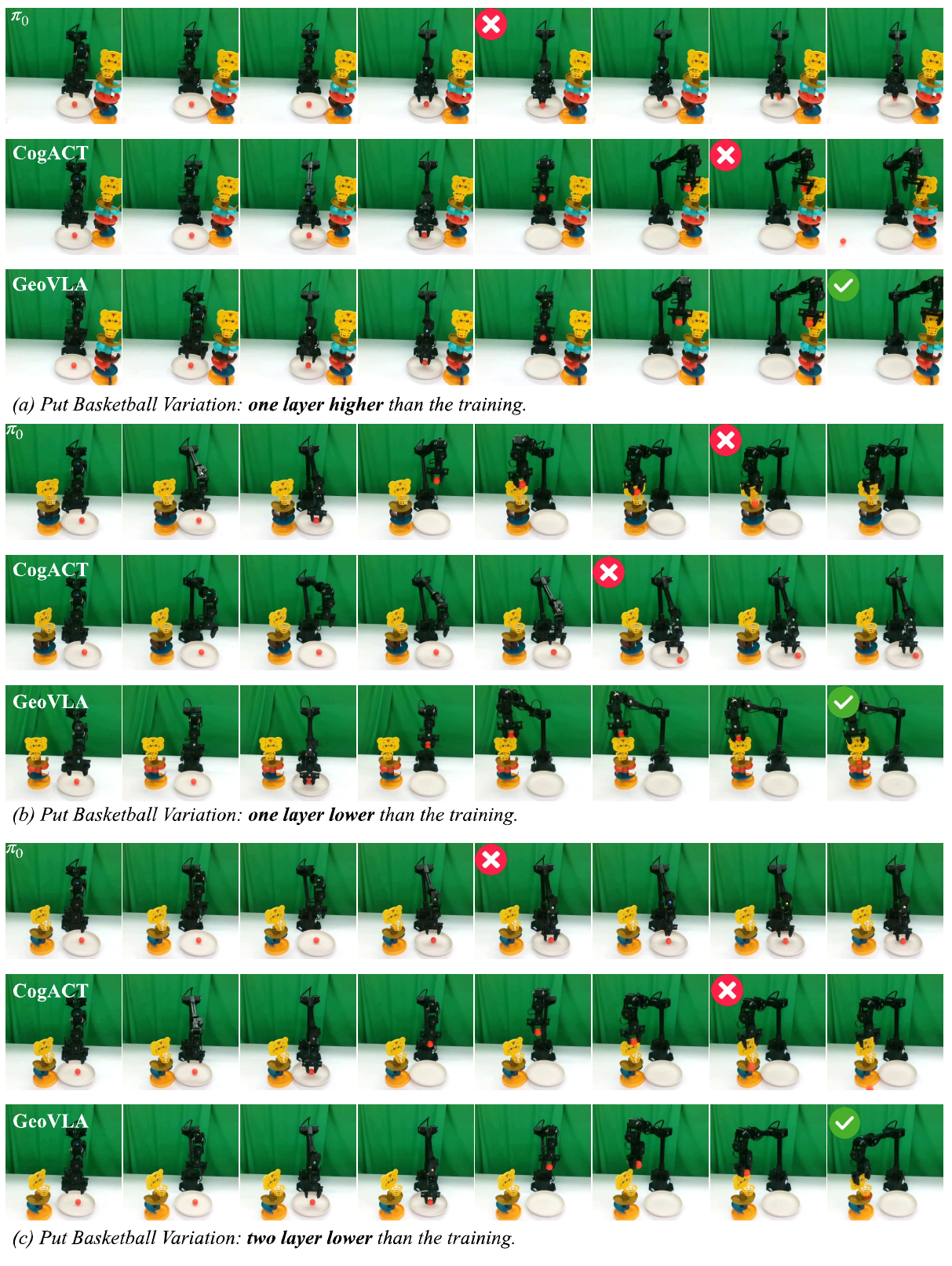}
\caption{\textbf{Qualitative comparison with $\pi_0$ and CogACT} on the \textit{Put Basketball} variation tasks. GeoVLA performs well on both higher and lower basketball position.}
\label{fig:07visual_variation}
\end{figure*}

\begin{figure*}[th]
\centering
\includegraphics[width=1.0\linewidth]{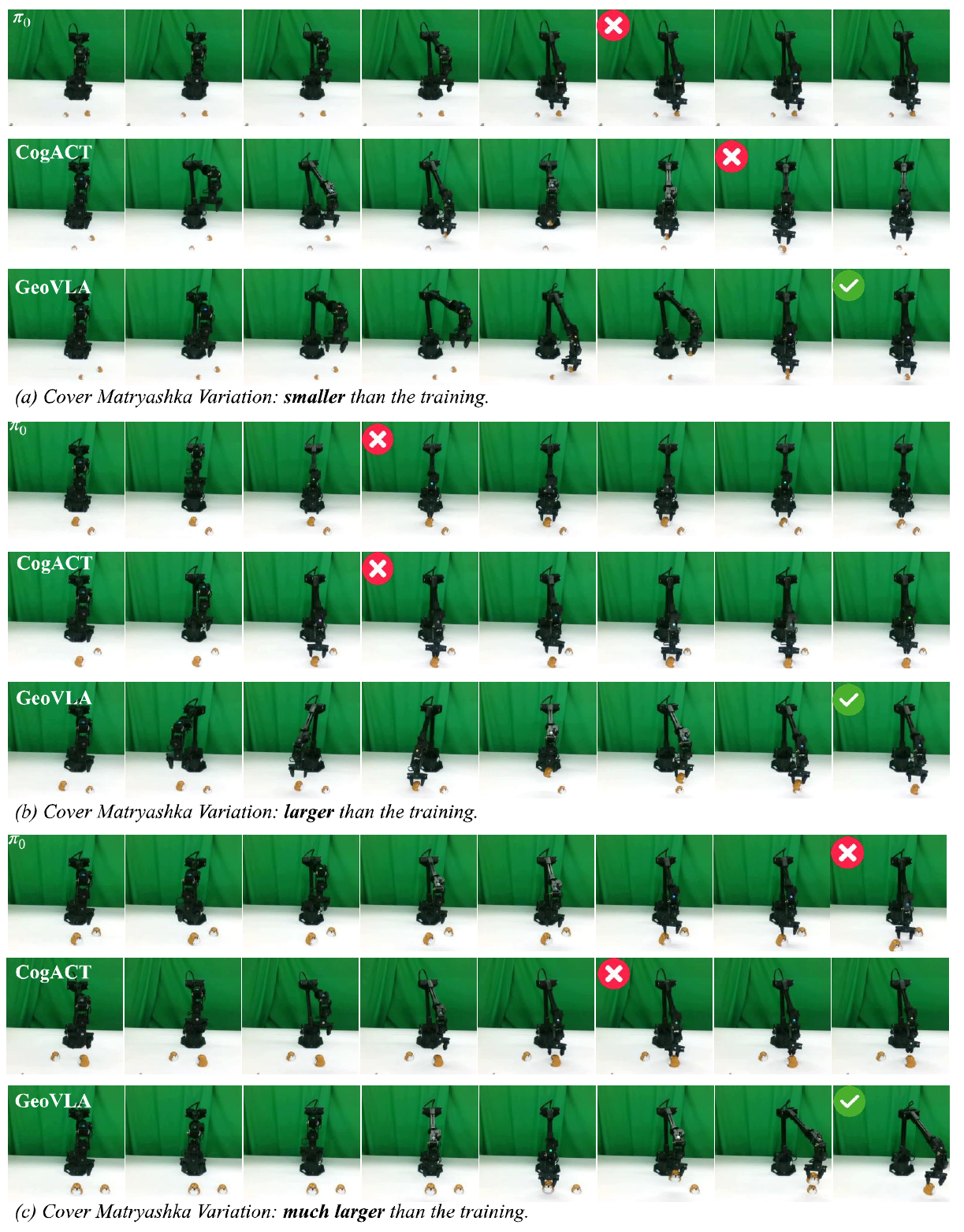}
\caption{\textbf{Qualitative comparison with $\pi_0$ and CogACT} on the \textit{Cover Matryashka} variation tasks. GeoVLA performs well on both larger and smaller doll sizes.}
\label{fig:07visual_variation_cover}
\end{figure*}

\begin{figure*}[th]
\centering
\includegraphics[width=1.0\linewidth]{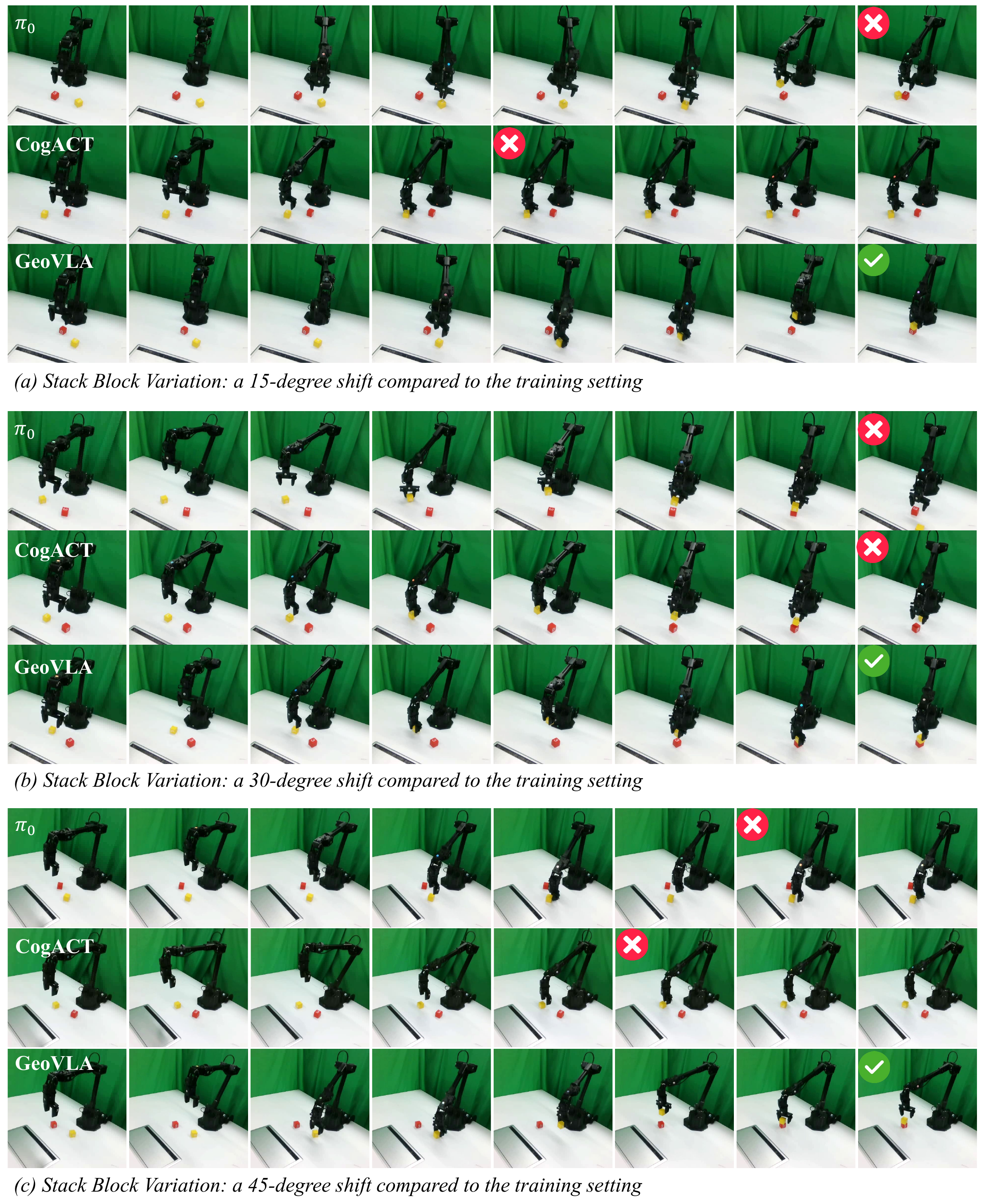}
\caption{\textbf{Qualitative comparison with $\pi_0$ and CogACT} on the \textit{Stack Block} variation tasks. GeoVLA performs better when the camera viewpoint changes.}
\label{fig:07visual_variation_shift}
\end{figure*}

In addition, Fig.~\ref{fig:07visual_variation} demonstrates that our proposed GeoVLA outperforms other SoTA methods such as $\pi_0$ (\cite{black2024pi_0}) and CogACT (\cite{li2024cogact}) by accurately recognizing the 3D positions in the \textit{Put Basketball} variation tasks.
Fig.~\ref{fig:07visual_variation_cover} shows that GeoVLA maintains strong performance when scaling the size of the dolls.
Fig.~\ref{fig:07visual_variation_shift} further illustrates the robustness of GeoVLA to camera viewpoint changes.

\section{More Experiments on Generalization}
\begin{figure*}[th]
\centering
\includegraphics[width=1.0\linewidth]{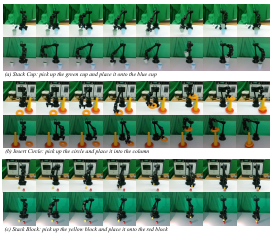}
\caption{\textbf{Visualization of GeoVLA on background and light change} on the \textit{Stack Cup}, \textit{Insert Circle}, \textit{Stack Block}. GeoVLA shows robustness to the background and light change.}
\label{fig:07visual_background_light}
\end{figure*}

To evaluate the robustness of the GeoVLA to the background and lightness, we conduct the experiments of changing the background and lightness on some basic tasks as shown in Tab.~\ref{tab:light_and_background}. Fig.~\ref{fig:07visual_background_light} also shows some qualitative visualizations.

\begin{table}[h!]
\centering
\caption{\textbf{Success rate (\%) under background and lighting variations} on three distinct manipulation tasks.}
\label{tab:light_and_background}
\begin{tabular}{l|ccc|c}
\toprule
Setting & Stack Cup & Insert Circle &Stack Block & Avg. \\
\midrule
w/o background change & 100 & 100 & 80  & 93.3\\
\midrule
w/ background change  & 90  & 70  & 80  & 80.0 \\
w/ light change       & 100  & 70  & 50 & 73.3\\
\bottomrule
\end{tabular}
\end{table}

% \subsection{Failure Cases}

\end{document}